\documentclass{article}

\PassOptionsToPackage{sort&compress}{natbib}
\usepackage[preprint]{corl_2026}

\usepackage[utf8]{inputenc} % allow utf-8 input
\usepackage[T1]{fontenc}    % use 8-bit T1 fonts
\usepackage{url}            % simple URL typesetting
\usepackage{booktabs}       % professional-quality tables
\usepackage{amsfonts}       % blackboard math symbols
\usepackage{amssymb}
\usepackage{nicefrac}       % compact symbols for 1/2, etc.
\usepackage{microtype}      % microtypography
\usepackage{xcolor}         % colors
\usepackage[table]{xcolor}
\usepackage{enumitem}
\usepackage{algorithm}
\usepackage{algpseudocode}
\usepackage{wrapfig}

\usepackage{graphicx}%
\usepackage{multirow}%
\usepackage{amsmath,amsfonts}%
\usepackage{amsthm}%
\usepackage{mathrsfs}%
\usepackage[title]{appendix}%
\usepackage{xcolor}%
\usepackage{textcomp}%
\usepackage{manyfoot}%
\usepackage{booktabs}%
\usepackage{algorithm}%
\usepackage{algorithmicx}%
\usepackage{algpseudocode}%
\usepackage{listings}%
\usepackage{printlen}
\usepackage{hyperref}
\usepackage{bm}
\usepackage{enumitem}
\usepackage{soul}
\usepackage{siunitx}
\usepackage{titlesec}

\titlespacing*{\section}
  {0pt}   % left indent
  {1.0ex} % space before
  {0.4ex} % space after

\titlespacing*{\subsection}
  {0pt}
  {0.8ex}
  {0.2ex}

\newcommand{\makesingular}[2]{\expandafter\newcommand\csname #1\endcsname{#2}}
\newcommand{\makeplural}[2]{\expandafter\newcommand\csname #1s\endcsname{#2s}}

\newcommand{\newWord}[2]{%
    \makesingular{#1}{#2}
    \makeplural{#1}{#2}
}

\newWord{methodName}{Neural Jacobian Field}
\renewcommand{\paragraph}[1]{\textbf{#1}}

\providecommand{\msdelete}[1]{}

\newcommand{\robotconfig}[1]{\mathbf{q}}
\newcommand{\robotcommand}[1]{\mathbf{u}}

\newcommand{\imageInput}[1]{\mathbf{I}}

\newcommand{\joints}{\ensuremath{\textbf{u}}}

\newcommand{\du}{\ensuremath{\delta \joints}}

\newcommand{\RR}{\mathbb{R}}

\DeclareMathOperator*{\argmin}{arg\,min}

\usepackage{color-edits}

\usepackage{tabularx}
\usepackage{array}
\usepackage{makecell}
\usepackage{pifont}

\usepackage{xspace}
\newcommand{\modelname}{Video-to-Embodied Robot Action Model\xspace}
\newcommand{\modelabbr}{\textsc{VERA}\xspace}
\newcommand{\jidm}{Jacobian IDM\xspace}
\newcommand{\includegraphicsfallback}[2][]{%
  \IfFileExists{#2}{\includegraphics[#1]{#2}}{%
    \fbox{\parbox[c][0.5\textwidth][c]{0.95\textwidth}{\centering Missing figure: \texttt{\detokenize{#2}}}}%
  }%
}

\newcommand{\tokenc}{E_{\mathrm{tok}}}
\newcommand{\tokdec}{D_{\mathrm{tok}}}

\addauthor{ms}{magenta}

\title{Turning Video Models into Generalist Robot Policies}

\author{
  \normalfont
  Sizhe Lester Li$^{1,}$\thanks{Equal contribution.} \quad
  Evan Kim$^{1,}$\footnotemark[1] \quad
  Xingjian Bai$^{1,}$\footnotemark[1] \\
  Tong Zhao$^{2}$ \quad
  Tao Pang$^{2}$ \quad
  Max Simchowitz$^{3,4}$ \quad
  Vincent Sitzmann$^{1}$ \\[4pt]
  $^{1}$MIT \quad
  $^{2}$Independent Researcher \quad
  $^{3}$CMU \quad
  $^{4}$Amazon FAR
}

\begin{document}

\raggedbottom % don't stretch white space to fill page bottoms

\setlength{\abovedisplayskip}{3pt}
\setlength{\belowdisplayskip}{3pt}
\setlength{\abovedisplayshortskip}{1pt}
\setlength{\belowdisplayshortskip}{1pt}

\maketitle

% CoRL requires two or three meaningful keywords.

%===============================================================================

% \vspace{-5mm}
\begin{figure}[!h]
  \centering
  \includegraphics[width=1.0\linewidth]{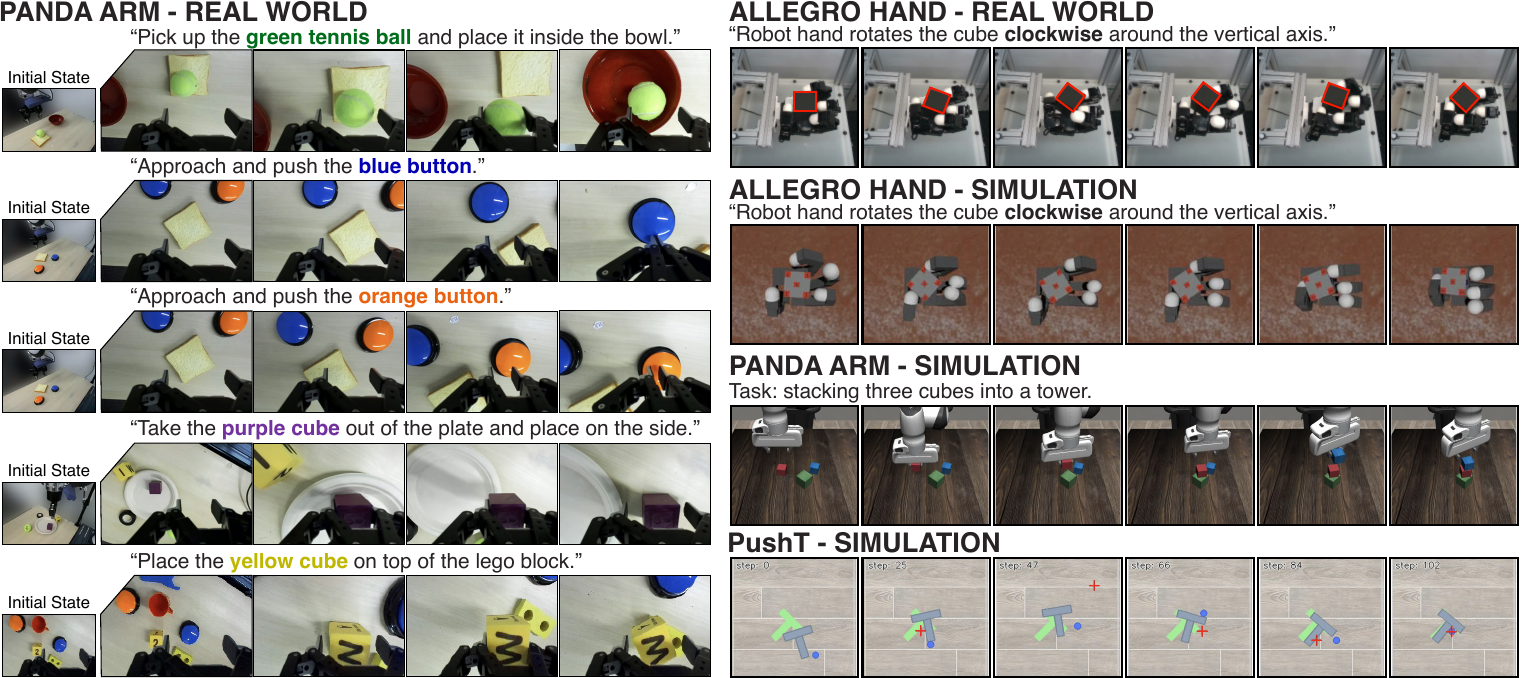}
  \vspace{-1.5em}
  \caption{\textbf{Controlling robots across embodiments, skills, and environments.}
    We demonstrate that a 14B video generative model, paired with an inverse dynamics model carefully designed for action-following, can solve a wide variety of robotic tasks, ranging from zero-shot pick-and-place tasks on a real-world Panda arm to contact-rich re-orientation of a cube with a 16-DoF robotic hand.}
  \label{fig:main-results}
  % \vspace{-2mm}
\end{figure}
\begin{abstract}
Video generative models have emerged as a promising robotics backbone, capable of generating videos that depict the completion of complex tasks across embodiments and environments.
Recent work proposes robot foundation models that jointly predict future observations and actions by finetuning video models with action-labeled data.
In this paper, we test the limits of an alternative approach: leave the video planner as-is while training an embodiment-specific inverse dynamics model (IDM). 
This decoupling offers several natural benefits: the video planner remains embodiment-agnostic, different video models can be interchanged easily without re-training the IDM, and the IDM can be independently trained with readily available self-play data.
We present a closed-loop, video-to-action policy that combines an action-free video world model with a carefully-designed IDM based on the robot embodiment Jacobian. 
We demonstrate that our IDM design is both data-efficient and scalable to high-dimensional action spaces.
Our policy, which we coin the \modelname (\modelabbr), achieves strong performance across simulated and real-world benchmarks, including zero-shot Panda arm manipulation and 16-DoF Allegro-hand dexterous cube re-orientation.
The same video planner can be used across multiple embodiments by pairing it with different embodiment-specific IDMs.
Our results show that decoupled video planning plus faithful video-to-action translation is a viable alternative route towards zero-shot, cross-embodiment, and generalizable robot control. More results are available on our project website~\footnote{{\textbf{Project Website:} \hypersetup{urlcolor=magenta}%
% --- ANONYMIZED (CoRL submission): comment OUT the line below and comment IN the anonymous one ---
\url{https://vera.csail.mit.edu}%
% \url{https://anonymous.4open.science/w/vjam-website-7AAE/}%
}}.

\end{abstract}

\section{Introduction}
\label{sec:intro}
The core challenge of robot intelligence is one of generalization: we seek a system that can control a wide variety of embodiments to solve unseen tasks in new environments.
% The core challenge of robot intelligence is generalization. A useful robot
% foundation model should
% transfer semantic knowledge, spatial reasoning, and physical priors across tasks,
% objects, scenes, camera configurations, and embodiments~\citep{10.5555/93002.93305, 10.5555/1121596, garrett2021tamp}. We seek a system that can control a wide variety of embodiments to solve unseen tasks in new environments.

In natural language processing, the generality of large language models (LLMs) and vision-language models (VLMs) emerged through jointly scaling models and datasets. Vision-language-action (VLA) models extend pretrained vision-language backbones with action outputs in an attempt to transfer their generality to robotics tasks~\citep{zitkovich2023rt2,kim2024openvla,octo2024,black2024pi0,pmlr-v305-black25a}.
However, different from the success of fine-tuning LLMs to VLMs, where web-scale, paired text and image data was available, no web-scale robot action data exists. In practice, it has proven challenging to achieve strong levels of generalization to new tasks, embodiments, and environments with VLAs~\cite{wang2025evaluating}: robot actions appear to be too distinct to inherit the generality of text and image models through fine-tuning on comparatively small action datasets.

Video generative models have been proposed as an alternative robotics backbone. They can generate ``visual plans'', imagining what it would look like for a robot to solve a specified task. 
Video models are not intrinsically tied to the action space of any one robot, and video data is abundant.
However, we must now solve the problem of translating the video plan into robot actions.
One route is to integrate actions into the video backbone itself, either through joint video-and-action prediction, known as \emph{world-action models} (WAMs), or through action-conditioned world models paired with inference-time search~\citep{gr2_2024,vpp2024,lingbot_va2026,ye2026worldactionmodelszeroshot,yang2024unisim,zhou2024dinowm,li2025unified}.
This route has produced impressive results, but, as with VLAs, requires a particularly rare kind of data for training: paired video, action, and text data.

In this work, we test the limits of an alternative approach to leveraging video models for robotics: a separate \textit{inverse dynamics model} (IDM) that takes video as input and outputs corresponding robot actions.
This separation has several key advantages. First, an IDM can be trained with \emph{task-agnostic} video-action data, which can be more readily generated e.g. via self-play. Second, the \textit{same} generative model can be shared across multiple embodiments by swapping the IDM. 
Lastly, stronger video backbones can be swapped in without retraining the IDM. 
However, the performance of the end-to-end pipeline is now critically dependent on the performance of the IDM: An inaccurate IDM leads to failures even if the video plan was perfect.
Prior IDMs regress actions directly from a pair of generated frames with an unconstrained network~\citep{du2023learning,ko2023avdc,rhoda2026dva}.
We find that, with limited action data, IDMs often generalize poorly out of distribution. Moreover, their accuracy degrades substantially as the complexity of the action space grows with more complex embodiments~(Sec.~\ref{sec:experiments}). 

To overcome these challenges, we introduce the \textit{Jacobian-IDM} (J-IDM), which predicts actions through the inversion of a learned tangent map between action perturbations and pixel motion. We show that such a structure is both data-efficient and scales gracefully with embodiment complexity. 
% Our work aims to demonstrate that, with better design, this approach \textit{can} be an alternative to other foundation model approaches. Specifically, we introduce the \textit{Jacobian-IDM} (J-IDM), which we show is both data efficient and scales gracefully with the complexity of the embodiment. 
We then pair J-IDM with a 14B video model adapted from the Large Video Planner~\citep{lvp2025}: 
the video model predicts a visual lookahead, J-IDM translates this lookahead into an executable action chunk, the system executes and observes, and finally replans based on the rollout. We call our closed-loop video-to-action policy \modelname (\modelabbr).

% Putting it together,  predicts a visual lookahead, a \textit{faithful} IDM converts this look-ahead into an action chunk for the current robot, 

\modelabbr is successful in simulated and real settings. On a real Panda arm, \modelabbr performs zero-shot language-conditioned manipulation in a new physical setup with varied camera placements, scene configurations, and prompts requiring visual and linguistic reasoning. On a real Allegro hand \citep{allegro}, \modelabbr performs RGB-only 16-DoF in-hand cube reorientation. In simulation, J-IDM outperforms prior IDM baselines.
Furthermore, we validate that a video planner trained across our real-world embodiments can be used with different embodiment-specific J-IDMs. Lastly, we situate \modelabbr against recent VLA and WAM baselines on real-world manipulation tasks. Our results show that video planning decoupled from a faithful inverse dynamics model is a promising and modular alternative for zero-shot, multi-embodiment, and generalizable robot control.

\section{Related Work and Preliminaries}
\label{sec:prelim}
\label{sec:preliminaries}

\newcommand{\jnn}{\mathbf J_\theta}
\newcommand{\tauvid}{\tau^{\mathrm{vid}}}
\newcommand{\tauact}{\tau^{\mathrm{a}}}
\newcommand{\bfs}{\mathbf{s}}
\newcommand{\bfa}{\mathbf a}
\newcommand{\bfo}{\mathbf{o}}
\newcommand{\bfg}{\mathbf{g}}
\newcommand{\dIDM}{D-IDM}
\newcommand{\jIDM}{J-IDM}
\newcommand{\VWM}{{{\pi^{vid}}}}

% \paragraph{Behavior Cloning and Vision-Language-Action Models.}
% %
% Cover RT-2, OpenVLA, Octo, $\pi_0$, $\pi_{0.5}$, GR-2, Diffusion Policy, ACT, BC-Z, Robomimic.
%

% \paragraph{Behavior Cloning and Vision-Language-Action Models.}
% Behavior cloning (BC) casts robot control as supervised action prediction from observations and task context, and remains a standard recipe for imitation-based manipulation~\citep{mandlekar2021robomimic,chi2023diffusion,zhao2023ACT}. Recent variants improve temporal expressivity by predicting action chunks or action distributions, as in ACT and Diffusion Policy~\citep{zhao2023ACT,chi2023diffusion}, and broaden task conditioning through language, goals, or videos, as in BC-Z~\citep{jang2022bcz}. Vision-language-action models scale this recipe by attaching robot actions to large visual/language backbones: RT-2 represents actions as language tokens~\citep{zitkovich2023rt2}, while OpenVLA, Octo, $\pi_0$, and $\pi_{0.5}$ train generalist policies over diverse robot data and action spaces~\citep{kim2024openvla,octo2024,black2024pi0, pmlr-v305-black25a}. These models inherit strong semantic priors from large-scale pretraining, but their control interface is still action-supervised and embodiment-specific. VJAM instead keeps the foundation model in video space and learns only a separate video-to-action bridge for each embodiment.

\paragraph{Behavior Cloning and Vision-Language-Action Models.}
Behavior cloning (BC) remains the dominant recipe for robot manipulation, learning actions directly from observations and task context, from offline visuomotor BC and language-conditioned imitation to more expressive sequence models based on action chunking, diffusion, and 3D action-centric representations~\citep{mandlekar2021robomimic,jang2022bcz,shridhar2023peract,jiang2023vima,zhao2023ACT,chi2023diffusion}. Modern VLA and generalist robot policies largely scale this same action-supervised interface: RT-1 and RT-2 attach robot actions to large visual/language backbones, while PaLM-E, RoboCat, Open X-Embodiment / RT-X, Octo, OpenVLA, and the $\pi$-family extend this paradigm to broader data mixtures, embodiments, and action spaces~\citep{brohan2022rt1,zitkovich2023rt2,driess2023palme,bousmalis2023robocat,openx2024rtx,octo2024,kim2024openvla,black2024pi0,pmlr-v305-black25a}. Despite their scale and generality across \textit{task} semantics, these methods still couple perception and control within a single embodiment-specific model. \modelabbr instead keeps the foundation model in video space and learns separate embodiment-specific video-to-action bridges for each embodiment.

% \paragraph{Video Generative Models.}
% %
% Large-scale video pretraining has produced increasingly capable generative models that exhibit emergent visual priors and rudimentary physical reasoning~\cite{sora2024} % paper: Video generation models as world simulators (Sora technical report), Brooks et al., OpenAI 2024
% ~\cite{yang2024cogvideox}. % paper: CogVideoX: Text-to-Video Diffusion Models with An Expert Transformer, Yang et al., arXiv 2024
% We build directly on the WAN family of open-weight video diffusion transformers~\cite{wan2025} % paper: Wan: Open and Advanced Large-Scale Video Generative Models, Wan Team (Alibaba), arXiv 2025
% as our backbone. For training, we adopt the Diffusion Forcing paradigm~\cite{chen2024diffusion}, % paper: Diffusion Forcing: Next-token Prediction Meets Full-Sequence Diffusion, Chen et al., NeurIPS 2024
% which assigns independent per-frame noise levels during training and thereby enables variable-context conditioning at inference. This per-frame independence is what we exploit to roll out arbitrary-length plans autoregressively from a clean context history, and recent extensions to transformer backbones~\cite{song2025dfot} % paper: History-Guided Video Diffusion (Diffusion Forcing Transformer), Song et al., 2025 \todo{verify exact title and venue}
% make the paradigm directly compatible with WAN-style architectures.

\paragraph{Video Generative Models.}
Large-scale video pretraining has produced increasingly capable generative models with emergent visual priors and rudimentary physical reasoning~\citep{sora2024,yang2024cogvideox}. We build on the WAN family of open-weight video diffusion transformers~\citep{wan2025} and train with Diffusion Forcing~\citep{chen2024diffusion,song2025dfot}, which enables variable-context conditioning. 

\paragraph{Joint Video-and-Action Models.}
A complementary line of work trains a single backbone to predict both future visual observations and robot actions, using video prediction as dense dynamics supervision while retaining an action head for execution. GR-2 jointly models video, language, and actions for manipulation~\citep{gr2_2024}; Video Prediction Policy learns predictive visual representations for policy learning~\citep{vpp2024}; and LingBot-VA and DreamZero / World Action Models co-predict world states and actions with a video model backbone~\citep{lingbot_va2026,ye2026worldactionmodelszeroshot}. Related action-conditioned world models and simulators use generated futures for planning or search~\citep{yang2024unisim,bruce2024genie,zhou2024dinowm}. These approaches couple visual forecasting and action generation, whereas \modelabbr separates them: the video model proposes an action-free visual plan, and an embodiment-specific IDM translates visual motion into controls.

\paragraph{Video Models with Inverse Dynamics Models.}
Most closely related to \modelabbr{} are methods that decouple robot control into a pixel-space planner and a translator~\citep{du2023learning,ko2023avdc,tian2025predictive,lvp2025,pai2025mimicvideo}. A similar factorization has also begun to appear in industrial systems: concurrent with this work, Rhoda AI's Direct Video-Action system achieves closed-loop, long-horizon bimanual manipulation in real-world settings~\citep{rhoda2026dva}. This result provides encouraging evidence that decoupled architectures can scale to real deployments.

% Rhoda AI's Direct Video-Action system pairs a causal video planner with a separate IDM, reporting closed-loop, long-horizon bimanual manipulation on demanding tasks such as multi-hour autonomous decanting and industrial container breakdown~\citep{rhoda2026dva}, providing encouraging evidence that this decoupling can scale to real deployments.
% While their progress provides encouraging industry-side signals that decoupling video planning from action translation can scale to real deployments, 
% it is communicated as a technical report without the model design and implementation details needed for reproduction and a scientific comparison.
% In this paper, we present a concrete design for a faithful IDM, together with the new capabilities that it unlocks.

\section{The \modelname}
\label{sec:method}

In this section, we first articulate the central technical challenge of using a video generative model as a robot policy --- translating predicted video into executable actions --- and identify the desired traits of a good translator (Sec.~\ref{sec:vid-to-act-problem}). We then introduce our proposed solution, the \jidm (Sec.~\ref{sec:jacobian}) and describe how we put it all together in a closed-loop with a video model in Sec.~\ref{sec:video-planner}.

% describe the video world model that serves as our visual planner (Sec.~\ref{sec:video-planner}), and finally explain how the two compose at deployment into a closed-loop receding-horizon controller (Sec.~\ref{sec:vja-policy}).

\newcommand{\pivid}{\pi_{\mathrm{vid}}}

\subsection{The Video-to-Action Problem}
\label{sec:vid-to-act-problem}

A video world model $\pivid$ produces a visual plan in pixel space: given an observation history $\bfo_{\le t}$ and a goal $g$, it samples
\begin{equation}
    (\hat{\bfo}_{t+1}, \hat{\bfo}_{t+2}, \ldots, \hat{\bfo}_{t+M})
\sim \pivid(\,\cdot\,\mid \bfo_{\le t}, g).
\end{equation}
Robots, however, do not act in pixels; they execute embodiment-specific commands such as position, velocity, or torque commands. The central question for the Video Model-IDM direction is therefore: \textit{how do we recover actions $\hat{\bfa}_{t:t+M-1}$ from video frames}?

\paragraph{Desidarata: Faithfulness, data-efficiency, and scaling with DoFs.}
A good IDM should, first and foremost, be \textit{faithful}: its inferred actions should reliably reproduce the predicted visual transition when executed. Beyond this, because action-labeled data is typically limited, it should be \textit{data-efficient}, best if it only requires self-play data to train. Finally, it should \textit{scale with action complexity}, allowing it to work for a variety of complex embodiments.

% remaining robust as the dimensionality and structure of the action space grow.

\paragraph{IDMs require careful designs.}
Current IDMs come in two flavors. The first is hand-crafted methods based on retargeting or 3D representations~\citep{lvp2025,ko2023avdc}. The second is direct parameterization via a neural network that learns to regress $\hat{\bfa}_t$ from the image pair $(\hat{\bfo}_t, \hat{\bfo}_{t+1})$ end-to-end~\citep{du2023learning}. Such an approach, which we call a \textit{Direct IDM} (\dIDM), has value in its simplicity. However, under limited data and complex embodiments, a more structured IDM may be needed to satisfy the previous criteria. In Section~\ref{sec:experiments} we show empirically that unstructured \dIDM s may sacrifice \emph{faithfulness} under (i) data constraints and (ii) increasing action dimension.

\subsection{The Jacobian Inverse Dynamics Model}
\label{sec:jacobian}
\label{sec:jacobian-idm}

%
% \paragraph{The embodiment Jacobian.}
% The problem of relating infinitesimal actions to motion is solved in classical robotics by the embodiment Jacobian. Let $\mathbf{q} \in \mathbb{R}^ {n_q}$ denote the robot's configuration or action vector and $\mathbf{x}_i(\mathbf{q}) \in \mathbb{R}^3$ the 3D location of body point $i$. The embodiment Jacobian
% \begin{equation}
% \label{eq:jacobian}
% \mathbf{J}_i(\mathbf{q}) \;=\; \frac{\partial \mathbf{x}_i(\mathbf{q})}{\partial \mathbf{q}} \;\in\; \mathbb{R}^{3 \times n_q}
% \end{equation}
% is the local linear map from action perturbations to point motion: $\delta \mathbf{x}_i \approx \mathbf{J}_i(\mathbf{q})\, \delta \mathbf{q}$. If one had access to the ground-truth Jacobian, then the motion of scene points observed in video could be combined with this local linear map to recover the underlying action by inversion.
% \citet{lesterli2025unifyingrepresentationcontrol} recently showed that one can learn to predict the Jacobian from training on limited self-play data. This satisfies our desired quality of data-efficiency. We therefore build on this insight, simplify the parameterization, and interpret the resulting model as a structured IDM.

\paragraph{The embodiment Jacobian.}
The problem of relating infinitesimal actions to motion is solved in classical robotics by the embodiment Jacobian. Let $\bfa \in \mathbb{R}^{n}$ denote the robot's action vector and $\mathbf{x}_i \in \mathbb{R}^3$ the 3D location of body point $i$. At a given state---observed as $\bfo$---the embodiment Jacobian
\begin{equation}
\label{eq:jacobian}
\mathbf{J}_i(\bfo) \;=\; \frac{\partial \mathbf{x}_i}{\partial \bfa}\bigg|_{\bfo} \;\in\; \mathbb{R}^{3 \times n}
\end{equation}
(Eq.~\ref{eq:jacobian}) is the local linear map from action perturbations to 3D point motion: $\delta \mathbf{x}_i \approx \mathbf{J}_i(\bfo)\, \delta \bfa$. However, 3D points are never observed directly, making such a Jacobian difficult to learn directly.
% We do not observe 3D points directly; what we observe is the image $\bfo$, and an action-induced change $\delta\bfo$ is captured per pixel by the projected 2D motion $\delta\mathbf{p}$ of the underlying body point. If one had access to the ground-truth Jacobian, then this observable pixel motion could be combined with the local linear map to recover the underlying action by inversion.

% \citet{lesterli2025unifyingrepresentationcontrol} previously parameterized this Jacobian as a \emph{3D} field, lifting per-pixel motion to 3D via volume rendering of a NeRF-style scene representation; we drop the 3D scene representation, which lets us scale $\jnn$ as a single, large image-conditioned transformer initialized from a generic vision backbone.

\paragraph{Image-space Jacobian field.}
This motivates predicting the Jacobian directly in image space, with $\bfo$ itself as the conditioning variable. Given a single image $\bfo \in \mathbb{R}^{H\times W \times 3}$, an image-conditioned transformer $\jnn$ outputs a dense field
\begin{equation}
    \jnn(\,\cdot\,,\bfo)
    \;:\;
    \{1,\dots,H\}\times\{1,\dots,W\}
    \;\longrightarrow\;
    \mathbb{R}^{2\times n},
\end{equation}
which assigns to every pixel $\mathbf{p}$ a $2{\times}n$ matrix that linearizes how an action increment $\delta\bfa\in\mathbb{R}^n$ moves that pixel:
\begin{equation}
    \delta \mathbf{p}
    \;=\;
    \jnn(\mathbf{p},\bfo)\,\delta\bfa,
    \label{eq:jacobian-flow}
\end{equation}
where $\delta\mathbf{p}$ is the per-pixel measurement of the observation change $\delta\bfo$, obtained in practice from an off-the-shelf optical-flow estimator (Eq.~\ref{eq:flow-loss}).
Prior work parameterized this Jacobian as a \emph{3D} field, lifting per-pixel motion to 3D via volume rendering of a NeRF-style scene representation~\cite{lesterli2025unifyingrepresentationcontrol}; we drop the 3D scene representation, which lets us scale $\jnn$ as a single, large image-conditioned transformer.
At test time, we recover the action by inverting Eq.~\ref{eq:jacobian-flow}.

\paragraph{Joint forward-inverse training objective.}
We train $\jnn$ on a dataset of $(\bfo_t,\,\delta\bfa_t,\,\bfo_{t+1})$ tuples. For each tuple, we extract a dense optical flow field $\mathbf{v}_t \in \mathbb{R}^{H\times W \times 2}$ between $\bfo_t$ and $\bfo_{t+1}$ using off-the-shelf motion estimators~\citep{teed2020raftrecurrentallpairsfield,doersch2023tapirtrackingpointperframe,harley2025alltracker,zhang2026megaflow}
% and treat $\mathbf{v}_t$ as a noisy measurement of $\delta\mathbf{p}$ at every pixel. 
. We supervise $\jnn$ with a joint forward-inverse loss combining the per-pixel Charbonnier objective $\rho(x){=}\sqrt{x^2+\varepsilon^2}$ on the predicted pixel motion with an action-reconstruction term using a $\lambda$-regularized pseudoinverse $\jnn^{\dagger,\lambda}$:
\begin{equation}
    \mathcal{L} \;=\;
    \underbrace{\sum_{\mathbf{p}} \rho\!\left(\jnn(\mathbf{p},\bfo_t)\,\delta\bfa_t - \mathbf{v}_t(\mathbf{p})\right)}_{\text{forward (pixel-flow)}}
    \;+\;
    w_{\text{a}} \underbrace{\sum_{\mathbf{p}} \big\|\delta\bfa_t - \jnn^{\dagger,\lambda}(\mathbf{p}, \bfo_t)\,\mathbf{v}_t(\mathbf{p})\big\|_2^2}_{\text{inverse (action reconstruction)}},
    \label{eq:flow-loss}
\end{equation}
with $w_{\text{a}}=0.3$; see App.~\ref{app:jac-train} for training details.

\paragraph{From forward model to IDM.}
At inference, we are given two consecutive predicted frames
$(\hat{\bfo}_t, \hat{\bfo}_{t+1})$ from the video planner. We extract
their optical flow $\mathbf v_t$ with the same off-the-shelf motion
estimator used in training and recover the action via the
$\lambda$-regularized pseudoinverse:
\begin{equation}
    \widehat{\delta \bfa}_t
    \;=\;
    \jnn^{\dagger,\lambda}(\hat{\bfo}_t)\,\mathbf v_t
    \;\triangleq\;
    \argmin_{\delta \bfa \in \RR^n}
    \sum_{\mathbf p}
    \big\|
        \jnn(\mathbf p, \hat{\bfo}_t)\,\delta \bfa
        -
        \mathbf v_t(\mathbf p)
    \big\|_2^2
    +
    \lambda \|\delta \bfa\|_2^2 .
    \label{eq:lstsq}
\end{equation}

% Two properties of this formulation are worth highlighting against the desiderata of Section~\ref{sec:vid-to-act-problem}. First, the action enters \emph{linearly}: $\jnn$ is forced to learn a structured tangent map rather than a free-form image-pair-to-action regressor, yielding both \emph{faithful} video-to-action translation and graceful scaling with action dimension. Second, $\jnn$ depends only on the embodiment, not on the video planner; switching planners or post-training data leaves the IDM untouched, preserving \emph{embodiment modularity}. Together with the self-play training pipeline above --- which addresses \emph{data efficiency} --- the construction meets all four properties laid out at the start of this section.

% Two properties of this formulation are worth highlighting against the desiderata of Section~\ref{sec:vid-to-act-problem}. First, the action enters \emph{linearly}: $\jnn$ is forced to learn a structured tangent map rather than a free-form image-pair-to-action regressor, addressing (D1) and (D4). Second, $\jnn$ depends only on the embodiment, not on the video planner; switching planners or post-training data leaves the IDM untouched, satisfying (D2). Combined with the self-play training pipeline above, the construction satisfies all four desiderata.

\subsection{Video World Model as a Closed-Loop Robot Planner}
\label{sec:video-planner}

\begin{figure}[!t]
  \centering
  \vspace{-2em}
  \includegraphics[width=\linewidth]{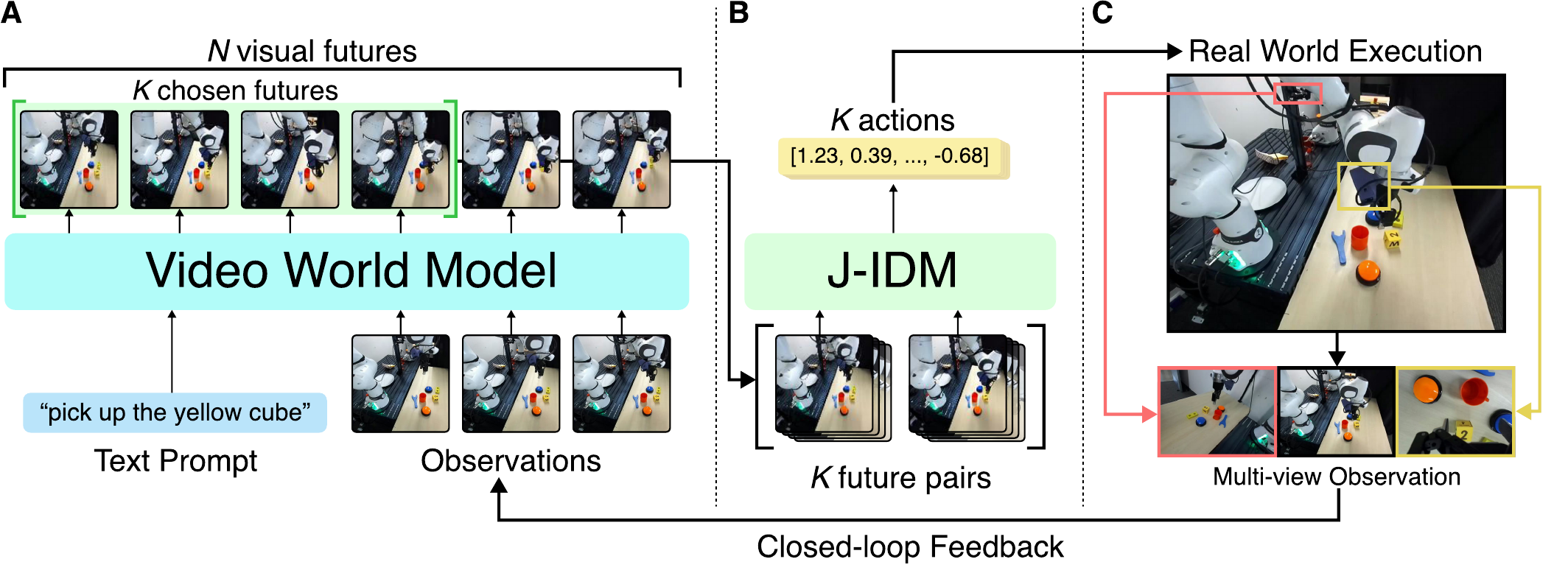}
  \vspace{-2em}
  \caption{\textbf{Translating video to actions.}
    \textbf{a,} Given context frames, our video world model rolls out a short visual plan.
    \textbf{b,} The Jacobian IDM inverts each step of this path into a chunk of low-level actions.
    \textbf{c,} The chunk is executed and observations return to the video model for closed-loop execution.}
  \label{fig:method-overview}
  \vspace{-0.5em}
\end{figure}

\modelabbr uses a video world model as a planner in observation space, as illustrated in Fig.~\ref{fig:method-overview}. Given some observations and conditioning signal (typically text), it predicts how the scene should visually evolve. Two design choices distinguish our planner from a vanilla video model.

\paragraph{Action-free robot post-training.}
We instantiate the planner from a pretrained video model~\citep{wan2025,lvp2025} and lightly adapt it to robot video-only data~\citep{khazatsky2024droid}; the post-training objective is generative video prediction with \emph{no action head}. Two modifications make it usable as a policy: video-to-video finetuning ($N$ context frames in, $M$ future frames out) for autoregressive rollouts, and a multi-view variant that tiles cameras into a single canvas. Architectural details are in App.~\ref{app:training}--\ref{app:video-model-ablations}.

% At deployment, \modelabbr runs as a receding-horizon controller. Each control update samples a visual look-ahead from the planner, converts a prefix of that look-ahead into actions with the \jidm, executes those actions on the robot, and replans from the resulting observations.

\paragraph{Executing on a visual chunk.}
Although the planner generates $M$ future frames, the controller commits to executing only the first $K$ predicted frames. The \jidm converts this committed prefix into a chunk of robot actions by applying Eq.~\ref{eq:lstsq} independently to every pair of adjacent frames, using the current observation $\bfo_t$ as the anchor:
\[
    \mathbf{\hat{a}}_{t:t+K}
    \;=\;
    G_\phi\bigl(\bfo_t,\hat{\bfo}_{t+1},\ldots,\hat{\bfo}_{t+K}\bigr),
\]
yielding a chunk length of $K$ actions. This allows the planner to reason over a longer visual look-ahead while the controller stays grounded through frequent feedback. Consistent with prior findings on action chunking, we observe that this chunking improves performance~\citep{zhao2023ACT,zhang2025action}; we ablate the choice of $K$ in Fig.~\ref{fig:action-chunking}.

\paragraph{Closed-loop replanning.}
After executing the chunk, the robot appends the newly observed frames to its history and queries the planner again. The execution horizon $K$ governs the feedback--smoothness trade-off: larger values produce longer locally-coherent chunks and amortize video generation time; smaller values reduce drift during execution. The full procedure is given in Alg.~\ref{alg:vja}.
\section{Experiments}
\label{sec:experiments}

\begin{figure}[!t]
  \centering
  \includegraphics[width=1.0\linewidth]{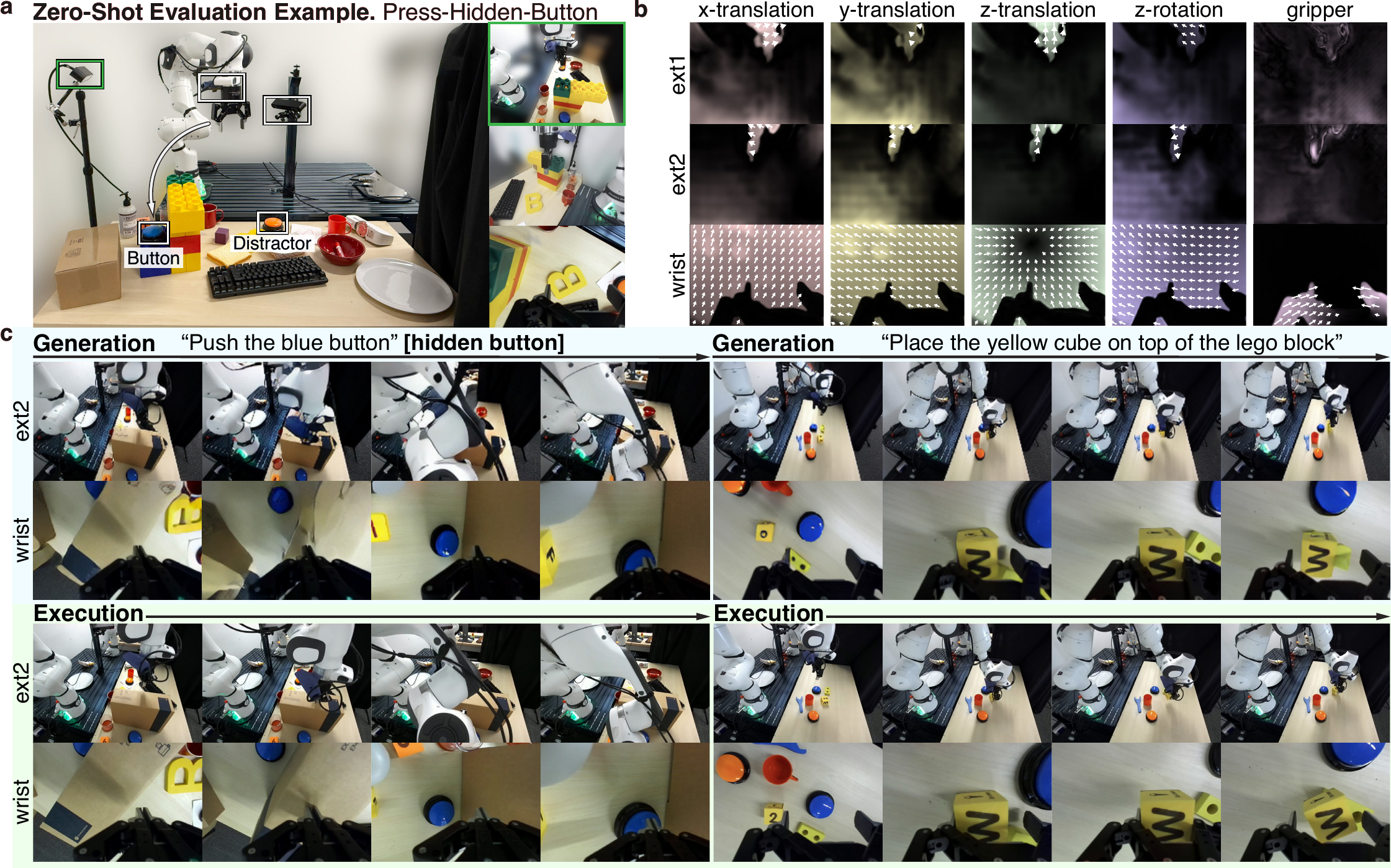}
  \vspace{-1.5em}
  \caption{\textbf{Zero-Shot Evaluation on Panda Arm.}
  \textbf{a,} After training on DROID~\cite{khazatsky2024droid}, we deploy zero-shot on a Panda in an unseen scene with ad-hoc camera placements; in \textit{Press-Hidden-Button}, the blue target is visible from only one of three views, with an orange distractor probing instruction grounding.
  \textbf{b,} Visualization of the predicted Jacobian: each column of $\jnn$ is assigned a fixed RGB color and rendered, per pixel, weighted by the norm of the corresponding Jacobian column. Columns: action channels; rows: viewpoints.
  \textbf{c,} Generated frames and executions shown side-by-side with the text prompt; actions remain closely aligned with the generated frames.}
  \label{fig:panda-zero-shot}
\end{figure}

We evaluate the \modelabbr{} system in closed-loop manipulation, isolate the \jidm{} against alternative IDM variants, and compare to state-of-the-art VLA and world-action baselines~\citep{ye2026worldactionmodelszeroshot,pmlr-v305-black25a}.

\paragraph{Experiment Setup.}
Unless otherwise noted, \modelabbr{} uses a Wan-family pretrained video model~\citep{wan2025} as the planner and an embodiment-specific \jIDM{} initialized from VGGT~\citep{wang2025vggt} as the translator. We evaluate in simulation (PushT, MimicGen Panda~\citep{mandlekar2023mimicgen}, Allegro-Sim cube reorientation~\citep{suh2025ctr}) and on hardware (Panda-Real post-trained on DROID~\citep{khazatsky2024droid}, Allegro-Real dexterous reorientation), reporting closed-loop success and task progress; full setup and video-model ablations are in Appendix~\ref{app:implementation}--\ref{app:video-model-ablations}.

\subsection{Main Results}
\label{subsec:main-res}

% We evaluate our approach across seven environments spanning simulation and the real world. In the real world, we conduct experiments on the panda arm and the allegro hand environment. In simulation, we evaluate our approach and baselines across the simulated panda arm environment \citep{mandlekar2023mimicgen}, the simulated allergro hand environment \citep{suh2025ctr}, and the PushT environment.

% Stuff we need to convey
% \begin{itemize}
%     \item faithful translation brings video model capabilities
%     \item those capabilities are: prompt following, multiview, cross-embodiment, visual reasoning
%     \item also zero-shot
%     \item allegro hand contact rich re-orientation
% \end{itemize}

\begin{figure}[!t]
  \centering
  \includegraphics[width=1.00\linewidth]{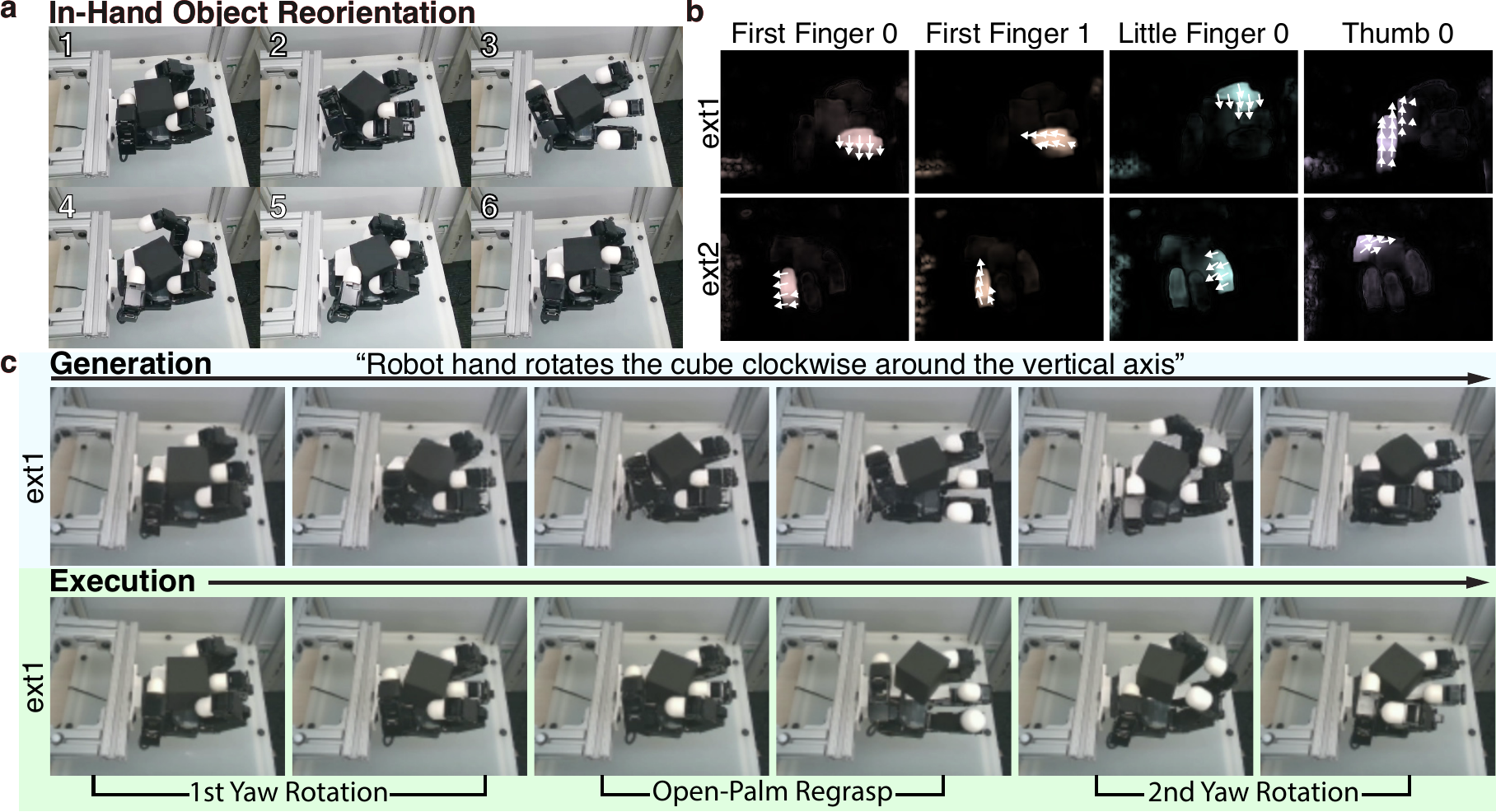}
  \caption{\textbf{In-Hand Object Reorientation.}
  \textbf{a,} Cube-reorientation demonstrations~\cite{suh2025ctr} with three language instructions (clockwise, counter-clockwise, random); we evaluate prompt-conditioned in-hand reorientation.
  \textbf{b,} Predicted Jacobian visualized as in Fig.~\ref{fig:panda-zero-shot}b (columns: selected action channels out of 16 total; rows: viewpoints).
  \textbf{c,} Generated frames and executions side-by-side with prompt; the model orchestrates dexterous finger motions.}
  \label{fig:allegro-inhand}
\end{figure}

\paragraph{Jacobian IDMs enable faithful video-to-action translation.}
We found that our \jidm can faithfully translate video chunks predicted by a world model into action chunks executed on the robot. Qualitatively, across environments, we have executions that are consistent with our generations, as shown in Fig.~\ref{fig:panda-zero-shot},\ref{fig:allegro-inhand}. Faithful translation naturally provides the robotic system with the capabilities of video models: visuospatial reasoning, prompt-following, embodiment generalization, and task generalization. These will be discussed throughout the rest of this section.

\paragraph{\modelabbr performs zero-shot, prompt-following manipulation tasks on a Panda arm.}
Just as any video model can generate videos zero-shot in new environments, when armed with a translator, \modelabbr performs zero-shot manipulation. Throughout our evaluations, we varied lighting conditions, camera configurations, and object orientations and found it robust to such changes. Our robot can perform manipulations that follow complex prompts, as demonstrated in Fig.~\ref{fig:main-results}, and perform complex vision-language reasoning tasks, as will be expanded upon in Sec.~\ref{subsec:foundation}.

\paragraph{\modelabbr admits multi-embodiment.}
Our video model which is finetuned across DROID and sim and real allegro hands --- two vastly different embodiments --- is able to generate successful video plans on both. Then, armed with two embodiment-specific \jIDM s, is able to control both a panda arm and the contact-rich re-orientation of a block via its fingers. As seen in Fig.~\ref{fig:allegro-inhand}, the latter is a multi-stage and challenging task for a purely visual system. However, armed with a strong video model, a sufficient translator, and closed loop control, \modelabbr is able to complete the task.

\subsection{Designing a Faithful IDM for Video-to-action Translation}
\begin{figure}[!t]
  \centering
  \includegraphics[width=\linewidth]{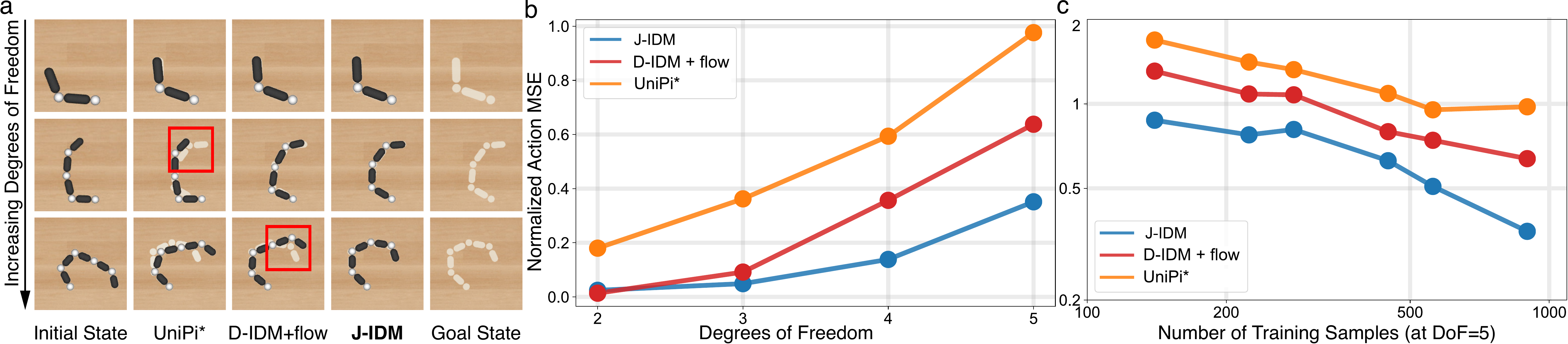}
  \vspace{-1.5em}
  \caption{\textbf{J-IDM scales with action dimensionality.}
    \textbf{a,} Rollouts of recovered actions for direct-IDM baselines and J-IDM as DoFs grow.
    \textbf{b,} At a fixed data budget, the J-IDM gap over direct-IDMs widens with DoF.
    \textbf{c,} At high DoF, J-IDM has a better data-accuracy trade-off across training sizes.}
  \label{fig:dof-scaling}
\end{figure}

\label{subsec:idm-ablations}

The closed-loop results above rely on \emph{faithful} video-to-action translation. We now isolate this property and ask what makes it possible. By comparing a Jacobian-parameterized IDM against unstructured \dIDM~baselines under controlled data and DoF budgets, we find the gain comes specifically from the Jacobian-based constraint. 
The \jIDM~is one concrete instantiation of that constraint; the rest of this subsection compares it against unstructured \dIDM~baselines.

% As established in Sec.~\ref{subsec:main-res} \jIDM 's faithful video-to-action translation unlocked the benefits of large-scale video models. In this subsection, we ablate the design choices behind our \jIDM~and compare against \dIDM s. 
% \evan{should we put some qualificiation note here about how it's not the end-all be-all IDM?}

\paragraph{Experiment Setup.} We compare our \jIDM~against blackbox \dIDM~approaches. In particular, we test against two models: one which takes in a pair of images (i.e. $\mathbf o_t$ and $\mathbf o_{t+1}$) and another which takes in an image and the optical flow between the two images. All models are kept at the same parameter count, trained on the same data, and share the same architecture with the exception of the decoding head. The flow-conditioned model serves as an ablation of our approach, since it has the same input-output behavior, but lacks our structured representation. The image-pair model is our best-effort replication of UniPi~\cite{du2023learning}. Because we use a different architecture to ensure a fair comparison, we refer to this baseline as UniPi*. More details can be found in Appendix~\ref{app:jac-ablate}.

\paragraph{\jIDM s are performant under data and complexity constraints.}
A controlled 2D ``toy finger'' study (Fig.~\ref{fig:dof-scaling}) sweeps degrees of freedom and self-play data quantity. The structured parameterization brings two benefits over an unstructured regressor: (1) at fixed data, \jIDM{} is the only model preserving faithful reconstruction as DoFs grow (Fig.~\ref{fig:dof-scaling}a,b); and (2) at fixed DoFs ($=$5), it is approximately $2\times$ more data-efficient (Fig.~\ref{fig:dof-scaling}c).

\paragraph{Faithful translation results in higher success rates.}
This fidelity carries downstream. In Tab.~\ref{tab:idm-ablation}, \jIDM{} achieves the lowest action-reconstruction MSE on all but one of the tested environments. Improved reconstruction translates to task success by preserving the video model's ``dreams'': with the planner held fixed, \jIDM{} outperforms UniPi* on every closed-loop task (Tab.~\ref{tab:sim-results}).

\begin{table}[!t]
  \centering
  % \vspace{1em}
  \caption{\textbf{Closed-loop results in simulated environments.}
    Each cell reports success rate / task progress (\%, $\uparrow$). \jIDM{} yields higher closed-loop success than a UniPi*-style direct inverse-dynamics baseline~\citep{du2023learning} across planar pushing, 7-DoF arm manipulation, and 16-DoF dexterous manipulation.}
  \label{tab:sim-results}
  \small
  \setlength{\tabcolsep}{5pt}
  \renewcommand{\arraystretch}{1.1}
  \begin{tabular}{lccc}
    \toprule
    Model                    & Allegro-Sim                              & Panda-Sim (MimicGen)                     & PushT-Sim                                \\
    \midrule
    Uni-Pi*                  & $0.0 \pm 0.0$\,/\,$0.0 \pm 0.0$          & $0.0 \pm 0.0$\,/\,$0.0 \pm 0.0$          & $74.4 \pm 3.4$\,/\,$84.8 \pm 2.8$        \\
    \jIDM{} \textbf{(ours)}  & $\mathbf{70.0 \pm 14.5}$\,/\,$70.0 \pm 14.5$ & $\mathbf{94.0 \pm 3.4}$\,/\,$94.0 \pm 3.4$ & $\mathbf{92.5 \pm 2.1}$\,/\,$95.5 \pm 1.6$ \\
    \bottomrule
  \end{tabular}
  % \vspace{-0.25em}
\end{table}

\begin{table}[!t]
  \centering
  \caption{\textbf{Action reconstruction for video-to-action translation.}
    Reconstruction MSE on held-out visual transitions. \jidm{} achieves the lowest MSE on Allegro-Sim, PushT-Sim, and 5-joint fingers, and remains competitive on Panda-Sim. \textbf{Best} / \underline{second-best}.}
    \vspace{-0.5em}
  \label{tab:idm-ablation}
  \small
  \setlength{\tabcolsep}{5pt}
  \renewcommand{\arraystretch}{1.1}
  \begin{tabular}{lcccc}
    \toprule
                              & \multicolumn{4}{c}{\textbf{Action reconstruction MSE} $\downarrow$}                                              \\
    \cmidrule(lr){2-5}
    \textbf{Model}            & Allegro-Sim        & Panda-Sim (MimicGen) & PushT-Sim          & 5-joint fingers    \\
    \midrule
    Uni-Pi*                   & 0.063              & 0.38                 & 0.071              & 0.047              \\
    \dIDM{} + Flow            & \underline{0.044}  & \textbf{0.09}        & \underline{0.059}  & \underline{0.030}  \\
    \jIDM{} \textbf{(ours)}   & \textbf{0.031}     & \underline{0.19}     & \textbf{0.046}     & \textbf{0.017}     \\
    \bottomrule
  \end{tabular}
  % \vspace{-0.01em}
\end{table}

%

% \evan{perhaps we want to swap the order of these two? should be easy to do}
% To diagnose the source of this advantage, we conducted a principled controlled study 

% \begin{wrapfigure}{hr}{0.5\textwidth}
% \centering
% \includegraphics[width=0.5\textwidth]{figures/Frame 2.pdf}
% % \caption{\textbf{Real-world manipulation results.}
% % %
% % We evaluate \modelabbr on two basic tasks (pick and push) and on more challenging tasks that require prompt, scene, and multi-view understanding. On basic tasks, \modelabbr performs on par with prior baselines, while on the challenging tasks it consistently outperforms them by a clear margin.
% % %
% % }
% % \caption{
% % \textbf{Real-world comparison on Panda manipulation.}
% % We compare \modelabbr{} with DreamZero and $\pi_{0.5}$ on basic pick-and-push tasks and more challenging tasks requiring multi-view, location-based, or semantic reasoning.
% % While DreamZero performs best on basic tasks, \modelabbr{} achieves higher success on the challenging tasks, where preserving the video planner's reasoning capabilities becomes more important.
% % }
% \caption{
% \textbf{Real-world comparison with robot foundation model baselines.}
% We evaluate \modelabbr{} against DreamZero and $\pi_{0.5}$ on two groups of Panda manipulation tasks.
% On basic tasks such as push and pick, \modelabbr{} remains competitive, while DreamZero performs best.
% On more challenging tasks that has occlusion, location-based reasoning, or semantic-based reasoning, \modelabbr{} achieves substantially higher success.
% }
% \label{fig:real-world}
% % \vspace{-4em}
% \end{wrapfigure}

\subsection{\modelabbr{} and other Robotic Foundation Models}
\label{subsec:foundation}

\begin{wrapfigure}[14]{r}{0.43\linewidth}
  \centering
  \vspace{-9pt}
  \includegraphics[width=\linewidth]{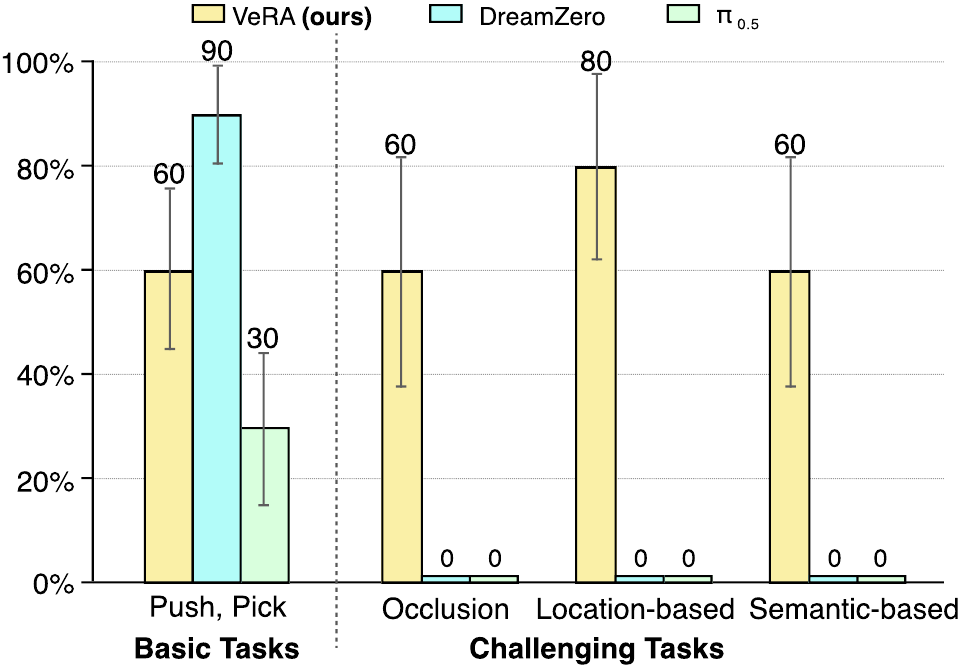}
  \vspace{-15pt}
  \caption{\footnotesize\textbf{Real-world comparison with robot foundation model baselines.}
    \modelabbr{} remains competitive on basic tasks and substantially better on challenging tasks.}
  \label{fig:real-world}
\end{wrapfigure}

We compare \modelabbr{} against state-of-the-art VLA and WAM baselines ($\pi_{0.5}$, DreamZero) on an in-house Panda manipulation suite, using DROID-trained checkpoints across all systems (Fig.~\ref{fig:real-world}).

\paragraph{Instruction following on basic tasks.}
On basic ``push A''/``pick up B'' tasks, DreamZero attains 90\%, \modelabbr{} 60\%, and $\pi_{0.5}$ 30\%. \modelabbr{}'s failures generally arise from the video-to-action step: \modelabbr{}'s dreamed futures on failing rollouts consistently complete the task, but the translation to action lacks fidelity. $\pi_{0.5}$, on the other hand, frequently disregards the instruction and acts on the wrong object.

\paragraph{Instruction following on challenging tasks.}
We further design two sets of harder, reasoning-heavy prompts: \textit{location-based} prompts (e.g., ``push the button on top of the paper'') that do not directly reveal the target, and \textit{semantic-based} prompts (e.g., ``push the button matching the wrench's color'') that require visual grounding. While video-to-action translation is the bottleneck for \modelabbr{} on basic tasks, the strength of the underlying video model allows \modelabbr{} to remain performant: DreamZero and $\pi_{0.5}$ frequently act on incorrect objects entirely, whereas \modelabbr{} succeeds, suggesting that decoupling the planner from the IDM preserves more of the video model's reasoning.

\paragraph{Reasoning through occlusions.}
To further test multi-view reasoning, we design the ``find the button'' challenge: a button is hidden behind a wall, visible from only one of three cameras, amid distractor props on a cluttered table (Figs.~\ref{fig:panda-zero-shot},~\ref{fig:hidden-button-appendix}). DreamZero and $\pi_{0.5}$ struggle to locate and navigate around the wall. \modelabbr{}, on the other hand, produces coherent plans that the J-IDM is able to execute. Moreover, for DreamZero, the dreamed future itself fails to produce a plan that finds and presses the button, indicating that the failure originates in the video branch. This is consistent with the hypothesis that preserving the video branch on its own allows the model to retain video-model reasoning capabilities that an end-to-end head would otherwise dilute.

\section{Discussion and Conclusion}
\enlargethispage{2\baselineskip}

\paragraph{Conclusion.}
We have shown that strong video models, paired with a faithful IDM, can control diverse embodiments and solve diverse tasks.
Our \jidm{} is one such instantiation: data-efficient, scaling with action dimensionality, and preserving the visual reasoning of the planner that an end-to-end action head would otherwise dilute.

\paragraph{Limitations.}
\modelabbr{} still requires robot-specific video for planner post-training, depends on off-the-shelf optical-flow trackers~\citep{teed2020raftrecurrentallpairsfield,harley2025alltracker} for \jIDM{} supervision, and cannot reason about force-based control~\citep{chen2022system} from RGB alone.
Faithful action recovery also degrades when the predicted transition has little observable pixel motion.

\paragraph{Future Work.}
Promising directions include sim-to-real \jIDM{} transfer to remove the need for real-robot action data, planner co-training across humans and robots, and an embodiment-conditioned Jacobian that serves multiple morphologies---motivating careful IDM design as video world models improve.

%===============================================================================
% \clearpage

% The acknowledgments are automatically included only in the final and preprint versions of the paper.
\acknowledgments{
This work was supported by the National Science Foundation under Grant No. 2211259, by the Intelligence Advanced Research Projects Activity (IARPA) via Department of Interior/ Interior Business Center (DOI/IBC) under 140D0423C0075, by the Amazon Science Hub, by the MIT-Google Program for Computing Innovation, by Advanced Micro Devices, Inc. under the AMD University Program's support of the MIT Hardware Consortium, and by a 2025 MIT Office of Research Computing and Data Seed Grant.
% If a paper is accepted, the final camera-ready version will (and probably should) include acknowledgments. All acknowledgments go at the end of the paper, including thanks to reviewers who gave useful comments, to colleagues who contributed to the ideas, and to funding agencies and corporate sponsors that provided financial support.
}

%===============================================================================

\bibliography{references}

\appendix
\section{Implementation Details}
\label{app:implementation}

This appendix collects implementation details omitted from the main exposition: video-model architecture and training recipe (\S\ref{app:training}), video-model ablations (\S\ref{app:video-model-ablations}), inference constants (\S\ref{app:inference}), datasets (\S\ref{app:datasets}), and tokenization and conditioning encoders (\S\ref{app:vae-encoders}). %, and reproducibility notes (\S\ref{app:repro}).
We use the notation from Sections~\ref{sec:preliminaries}--\ref{sec:method}: $N$ context frames, $M$ predicted frames, and a committed prefix of length $K$.

\subsection{Architecture and Training Recipe of Video Models}
\label{app:training}

\paragraph{Backbone family.}
We instantiate the video planner $F_\theta$ with the Wan family of open-weight video diffusion transformers~\citep{wan2025}.
The video planner is always trained and used in observation space: it predicts future visual observations from a short observation history and, when available, a task specification.
We do not attach an action head to the video model, and the video model training objective never uses robot actions.
This design keeps the planner compatible with action-free video pretraining and separates visual planning from embodiment-specific control.

\paragraph{Model variants.}
We use two model scales throughout the paper.
The Wan-1.4B model is used for development runs and video-model ablations, while the Wan-14B model is used for the main large-scale experiments.
The 14B runs are initialized from pretrained Wan-family checkpoints; in the main robot-control setting, we use the Large Video Planner (LVP) initialization~\citep{lvp2025} when specified.
For ablations, we also consider smaller models trained from different initializations to measure the effect of video pretraining.
Table~\ref{tab:video-model-variants} summarizes the stable configuration details.

\begin{table}[!htbp]
\centering
\small
\caption{\textbf{Video model variants.} We report only stable configuration details. The smaller model is used for development and ablations, while the 14B model is used for the main experiments.}
\label{tab:video-model-variants}
\begin{tabular}{llll}
\toprule
Variant & Role in this paper & Initialization & Training hardware \\
\midrule
Wan-1.3B & Development and ablations & Wan-family checkpoint & 1 H100 GPU \\
Wan-14B & Main experiments & Wan-family / LVP checkpoint & 1 H200 node, 8 GPUs \\
\bottomrule
\end{tabular}
\end{table}

\paragraph{Robot video post-training.}
For robot-control experiments, we post-train the planner on robot video data using the same generative prediction objective as in video pretraining.
The target remains future RGB observations rather than actions.
This post-training adapts the model to robot scenes, camera viewpoints, objects, and contact dynamics, while preserving the action-free interface of the planner.
Dataset-specific details, including camera views, resolutions, frame counts, and train/validation splits, are provided in Appendix~\ref{app:datasets}.

\paragraph{Video-to-video training format.}
Each training example consists of a context segment and a future segment.
Given $N_{\mathrm{ctx}}$ context latent frames, where $N_{\mathrm{ctx}} \leq N$, the model predicts a future look-ahead of $M$ latent frames.
The context frames are used as conditioning, and the training loss is applied only to the future frames.
We use this video-to-video format because it matches closed-loop deployment: at test time, the robot provides a recent observation history, and the planner proposes a short visual future to be translated into actions.

\paragraph{Diffusion-forcing objective.}
We train the planner with diffusion forcing~\citep{chen2024diffusion}, following the history-guided / DFoT-style formulation~\citep{song2025dfot}.
In diffusion forcing, different frames in a sequence can be assigned different noise levels, allowing the model to condition on clean or partially noised history frames while denoising future frames.
For a future latent frame $z_i$, we use the standard linear-flow diffusion target and optimize mean-squared error on the future segment:
\begin{equation}
    \mathcal{L}_{\mathrm{vid}}(\theta)
    =
    \mathbb{E}
    \left[
        \sum_{i=1}^{M}
        \left\|
            \hat{\mathbf{v}}_{\theta,i}
            -
            \mathbf{v}_{i}
        \right\|_2^2
    \right],
\end{equation}
where $\mathbf{v}_{i}$ is the diffusion target for the $i$-th future latent frame and $\hat{\mathbf{v}}_{\theta,i}$ is the model prediction.
Context frames are used for conditioning and are not included in this loss.

\paragraph{Context and look-ahead.}
The maximum context length $N$ and the future look-ahead length $M$ are treated as design choices.
Longer context can improve temporal consistency and help disambiguate contact state, while longer look-ahead gives the inverse dynamics module a richer visual path to follow.
However, longer sequences also increase memory and computation and can make generation harder.
We therefore use short local look-aheads in the closed-loop controller and replan frequently from new observations.
The ablations over context length, look-ahead length, and related planner choices are reported in Appendix~\ref{app:video-model-ablations}.

\paragraph{Multi-view formatting.}
When multiple camera views are available, we concatenate them in pixel space and treat the result as a single video frame.
This simple formatting lets the same backbone consume third-person and wrist-camera observations without architectural changes.
For cross-embodiment training, all embodiments are converted to a shared view layout; when an embodiment provides fewer views, the missing view slots are padded with blank frames.
This convention lets a single video planner be trained on a mixture of embodiments while leaving embodiment-specific action grounding to the inverse dynamics module.

\paragraph{Training hardware and implementation scope.}
The Wan-14B planners are trained on one node of 8 H200 GPUs.
The Wan-1.4B planners are trained on a single H100 GPU.
All runs use mixed-precision training when supported by the hardware and software stack.
We intentionally omit low-level optimizer settings, sharding policies, and batch-size details from this section because these are implementation-dependent and may vary across development runs.

\subsection{Video Model Ablations}
\label{app:video-model-ablations}

This section summarizes the video-planner ablations used for model selection and design validation.
All reported numbers are validation MSE losses on held-out future-frame prediction, computed under the diffusion-forcing objective described in Appendix~\ref{app:training}.
These losses are useful for comparing video-model configurations, but they are not a substitute for the closed-loop success metrics reported in the main experiments.
To avoid over-interpreting development diagnostics, we report only stable comparisons and round all validation losses to two decimal places.

\paragraph{Initialization and model scale.}
Table~\ref{tab:video-init-scale-ablation} compares the effect of video-model initialization and scale.
The randomly initialized model is trained only on robot videos, while the pretrained variants start from Wan-family or LVP checkpoints~\citep{wan2025,lvp2025}.
Rows marked with $\dagger$ terminated before their planned training budget and should be interpreted as development diagnostics rather than fully controlled scaling results.

\begin{table}[!htbp]
\centering
\small
\caption{\textbf{Video-model initialization and scale ablations.}
Validation MSE loss is measured on held-out future latent prediction and rounded to two decimal places.
}
\label{tab:video-init-scale-ablation}
\setlength{\tabcolsep}{4pt}
\begin{tabular}{llllccc}
\toprule
Name & Backbone & Init. & Dataset / views & $N$ & $M$ & Val. MSE $\downarrow$ \\
\midrule
Random init & Wan-14B & random & DROID triview & 8 & 6 & $0.40$ \\
Wan-14B & Wan-14B & Wan checkpoint~\citep{wan2025} & DROID triview & 8 & 6 & $0.13$ \\
LVP-14B & Wan-14B & LVP warm-start~\citep{lvp2025} & DROID triview & 8 & 6 & $0.10$ \\
\bottomrule
\end{tabular}
\end{table}

The comparison supports two qualitative conclusions.
First, pretrained video initialization substantially improves robot-video prediction: the Wan-initialized 14B model achieves much lower validation loss than the randomly initialized model under the same data and view configuration.
Second, the LVP-initialized 14B model gives the lowest validation loss among the configurations we report, consistent with the intuition that stronger video priors improve visual planning.

\paragraph{Context and look-ahead.}
We treat the context length $N$ and visual look-ahead length $M$ as deployment hyperparameters rather than reporting them as a standalone video-quality table.
Longer context can provide more temporal information and help disambiguate contact state, while longer look-ahead gives the inverse dynamics module a richer local visual path to follow.
At the same time, longer sequences increase memory and computation and can make future prediction harder.
For the main closed-loop controller, we therefore use short local look-aheads and rely on frequent replanning from new observations; the canonical inference quantities are summarized in Appendix~\ref{app:inference}.

\paragraph{Camera-view configuration.}
Table~\ref{tab:video-view-ablation} compares the camera views supplied to the DROID video planner.
The triview configuration concatenates two external views and a wrist view; the reduced-view configurations remove one or more of these streams.
Because the raw loss is computed on different image widths, we normalize losses to the triview pixel budget by multiplying the measured validation loss by $576/W$, where $W$ is the concatenated input width.

\begin{table}[!htbp]
\centering
\small
\caption{\textbf{Camera-view ablations.}
Validation MSE loss is measured on held-out DROID clips with the same view configuration used for training, normalized to the triview $128 \times 576$ pixel budget, and rounded to two decimal places.}
\label{tab:video-view-ablation}
\setlength{\tabcolsep}{4pt}
\begin{tabular}{lllccc}
\toprule
Name & Views & Resolution & $N$ & $M$ & Norm. Val. MSE $\downarrow$ \\
\midrule
Triview & ext1 + ext2 + wrist & $128 \times 576$ & 8 & 6 & $0.13$ \\
External + wrist & ext + wrist & $128 \times 384$ & 8 & 6 & $0.21$ \\
Two external views & ext1 + ext2 & $128 \times 384$ & 8 & 6 & $0.11$ \\
Single external view & ext1 & $128 \times 192$ & 8 & 6 & $0.21$ \\
Wrist-only & wrist & $128 \times 192$ & 8 & 6 & $0.61$ \\
\bottomrule
\end{tabular}
\end{table}

The view ablation suggests that global scene context is especially important for video prediction.
Using two external views gives the lowest normalized validation loss, likely because the model observes the object, robot, and workspace from stable third-person viewpoints.
The triview model is slightly worse by this metric but remains attractive for policy use because it combines global context with the wrist view, which can provide useful local information near contact.
Removing one external view degrades prediction quality, and the wrist-only model performs substantially worse, consistent with the difficulty of predicting the full scene from a narrow, moving camera.
We therefore use multi-view inputs in the main real-robot planner whenever available.

\subsection{Inference Constants and Closed-Loop Rollout}
\label{app:inference}

\paragraph{Video sampling.}
At each policy update, the planner encodes the most recent $N$ observations and samples $M$ future latent frames with the UniPC sampler used by the Wan backbone~\citep{wan2025}. The context latents are kept fixed and only the future latents are denoised. Unless otherwise stated, we use $40$ UniPC denoising steps per call.

\paragraph{Closed-loop execution.}
\modelabbr runs as a receding-horizon controller. At each step, the planner predicts $M$ future frames; the controller decodes the first $K$ and hands them to the \jidm, which produces $K$ actions by applying Eq.~\ref{eq:lstsq} to each adjacent pair (anchored at the current observation). After executing the chunk, the robot receives new observations and the planner is queried again. Frames beyond the committed prefix are discarded.

\paragraph{Canonical operating point.}
Table~\ref{tab:closed-loop-quants} lists the values used in the main experiments. We keep $K$ small so the robot replans often from fresh observations --- important for contact-rich manipulation.

\begin{table}[!htbp]
\centering
\small
\caption{\textbf{Canonical closed-loop inference constants.}}
\label{tab:closed-loop-quants}
\begin{tabular}{lll}
\toprule
Quantity & Symbol & Value \\
\midrule
Context length (latent frames) & $N$ & 6 \\
Predicted look-ahead (latent frames) & $M$ & 4 \\
Committed prefix (latent transitions) & $K$ & 1 \\
Video sampler steps & --- & 40 UniPC \\
\bottomrule
\end{tabular}
\end{table}

The number of low-level robot commands per committed latent transition depends on the task control rate and the video frame stride, and is set per environment.

\subsection{Datasets and Embodiments}
\label{app:datasets}

We evaluate \modelabbr{} across simulated and real-world manipulation settings spanning planar pushing, 7-DoF arm manipulation, and 16-DoF dexterous in-hand manipulation.
Table~\ref{tab:envs} summarizes the main dataset and embodiment statistics.
Across all settings, the video planner is trained on RGB observations only.
The \jidm{} additionally uses paired robot actions and visual motion estimates to learn embodiment-specific video-to-action translation.

\begin{table}[!htbp]
\centering
\footnotesize
\setlength{\tabcolsep}{3pt}
\renewcommand{\arraystretch}{1.08}
\caption{\textbf{Datasets and embodiments.}
``Resolution'' denotes the RGB input resolution after view concatenation or padding.
Episode counts refer to the main trajectories used for video-model post-training and/or \jidm{} training in each setting.
Real-world performance is evaluated by executing the learned policy on physical robots rather than by held-out trajectory prediction.}
\label{tab:envs}
\begin{tabularx}{\linewidth}{@{}l l c r c >{\raggedright\arraybackslash}X@{}}
\toprule
Environment & Type & \makecell{Robot\\DoF} & Episodes & Resolution & Cameras \\
\midrule
PushT-Sim & Sim & 2 & 206 & $128{\times}384$ & one top-down view, padded to a shared canvas \\
\makecell[l]{Panda-Sim\\(MimicGen)} & Sim & 7 & 9{,}000 & \makecell[c]{$128{\times}128$\\$128{\times}256$} & eye-in-hand, agent-view, dualview \\
Allegro-Sim & Sim & 16 & \makecell[r]{2{,}126\\(+401 aux.)} & $128{\times}384$ & triview \\
Panda-Real & Real & 7 & 39{,}816 & $128{\times}576$ & two external views + wrist view \\
Allegro-Real & Real & 16 & 60--217 & $128{\times}256$ & dualview \\
\bottomrule
\end{tabularx}
\end{table}

\paragraph{PushT-Sim.}
PushT-Sim is a planar pushing benchmark in which a circular 2-DoF pusher moves a T-shaped block toward a target pose.
We use the expert demonstration set with $206$ episodes.
The single top-down RGB view is resized to $128 \times 128$ and placed into a padded $128 \times 384$ canvas so that the video planner can share the same multi-view formatting convention used in other settings.
The action is represented as a normalized 2D pusher-position delta between adjacent observations.
For \jidm{} training, visual motion is estimated from RGB observations.
Success follows the standard PushT target-overlap criterion, and evaluation rollouts are capped at $200$ environment steps.

\paragraph{MimicGen (Panda-Sim).}
MimicGen (Panda-Sim) consists of simulated Franka Panda manipulation trajectories generated with the MimicGen pipeline~\citep{mandlekar2023mimicgen}.
We use four tasks: \texttt{square}, \texttt{coffee}, \texttt{stack}, and \texttt{stack\_three}, with $3{,}000$, $2{,}000$, $2{,}000$, and $2{,}000$ episodes respectively.
Each episode provides an external agent-view camera and an eye-in-hand camera.
We use $128 \times 128$ resolution for individual views and $128 \times 256$ for the concatenated dual-view input.
The action space is a 7-DoF end-effector command with gripper state.
Success is determined by the original MimicGen task specification.

\paragraph{Allegro-Sim.}
Allegro-Sim is a simulated cube reorientation setting with a 16-DoF Allegro hand.
The action is a 16D joint-position target tracked by a low-level controller.
The main task is relative cube reorientation, where the hand rotates the cube from its initial pose toward a target relative orientation.
We use $2{,}126$ planner-generated reorientation episodes for the main task.
We additionally use $401$ auxiliary random-motion episodes to provide local motion coverage for \jidm{} training; these auxiliary trajectories are not goal-directed demonstrations of the reorientation task.
Three calibrated RGB views are concatenated width-wise into a $128 \times 384$ input.
For \jidm{} training, we use paired visual observations, actions, and visual motion estimates from the same simulated trajectories.

\paragraph{Allegro-Real.}
Allegro-Real consists of real-world trajectories collected on a horizontally mounted Allegro hand.
We use a Phase~1 dataset of $60$ episodes and a larger Phase~2 pretraining mix of up to $217$ episodes.
The tasks involve clockwise and counter-clockwise cube reorientation.
Two calibrated RGB cameras are concatenated into a $128 \times 256$ dual-view input.
Evaluation is performed by executing the policy on the physical robot under held-out trial conditions.

\paragraph{Panda-Real.}
Panda-Real is our real-world Franka-family tabletop manipulation setting.
For video-planner post-training, we use the successful subset of DROID~\citep{khazatsky2024droid}, consisting of $39{,}816$ real-robot trajectories collected with Franka-family arms and parallel grippers.
We use two external cameras and one wrist camera, concatenated into a $128 \times 576$ RGB input.
The video planner is trained on RGB observations only.
For the Panda \jidm{}, we use the corresponding action streams in the Panda control parameterization and visual motion estimated from RGB observations.
Unless otherwise noted, Panda actions are represented as end-effector delta commands with gripper state.
Evaluation is conducted on our physical Panda setup with held-out prompts, object configurations, and camera placements.

\paragraph{Train / validation / evaluation protocol.}
For MimicGen, we hold out $250$ episodes uniformly across tasks for validation and use the remaining episodes for training.
Closed-loop simulation evaluation uses fresh rollout initializations or task instances rather than simply replaying held-out trajectories.
For real-world settings, evaluation is performed by executing the learned policy on the physical robot and reporting task success over held-out trials.
Across all environments, closed-loop task success is the primary evaluation metric; validation losses are used only for model selection and diagnostics.

\subsection{Tokenization and Conditioning Encoders}
\label{app:vae-encoders}

\paragraph{Video tokenizer and decoder.}
We use the pretrained video tokenizer and decoder from the Wan model family~\citep{wan2025}.
The tokenizer maps RGB video clips into latent video tokens, and the decoder maps generated latent tokens back into RGB frames.
Both components are kept fixed during the video model finetuning.
This keeps the video planner objective in the same latent video space as the pretrained backbone and avoids introducing any action-specific representations into the video model.

\paragraph{Temporal layout.}
The Wan tokenizer uses a causal temporal compression.
We denote by $N$ the number of context latent frames and by $M$ the number of predicted future latent frames.
For the clip lengths used in our experiments, $N$ latent context frames correspond to $4N{-}3$ RGB context frames, while $M$ predicted latent frames correspond to $4M$ future RGB frames.
These conversions are used only to describe the closed-loop operating point; the controller itself operates by encoding recent RGB observations, sampling future latent frames, decoding the committed prefix, and translating that visual prefix into actions with the \jidm{}.

\paragraph{Task conditioning.}
When task-conditioning signals are available, such as language prompts in real-world manipulation, the video planner conditions on them during video prediction.
When such signals are not available or not used, the same architecture is trained as a video-to-video predictor from observation history alone.
This allows the planner to support both prompt-conditioned real-robot settings and task-structured simulation settings without changing the video-to-action interface.

\paragraph{Multi-view formatting.}
For multi-camera settings, camera views are concatenated in pixel space before tokenization.
The tokenizer therefore receives a single wide RGB frame containing all available views.
This convention is simple but useful: it allows the same video backbone and tokenizer to handle single-view, dual-view, and triview inputs without architectural changes.
For cross-embodiment video training, missing view slots are padded with blank frames so that all embodiments can share a common input layout.

\paragraph{Separation from actions.}
Robot actions are not tokenized by the video model and are never predicted by the video planner.
Actions are used only to train the embodiment-specific \jidm{}.
At deployment, the planner produces future visual observations, the decoder converts the committed latent prefix into RGB frames, and the \jidm{} translates those frames into low-level robot actions.
This maintains the central factorization of \modelabbr{}: action-free visual planning in the video model and embodiment-specific control in the inverse dynamics module.

\begin{algorithm}[t]
  \caption{\modelabbr{} policy}
  \label{alg:vja}
  \begin{algorithmic}[1]
    \Require video world model $F_\theta$, \jidm{} $G_\phi$, tokenizer $\tokenc$, detokenizer $\tokdec$,
             conditioning signal $c$, context length $N$, look-ahead length $M$,
             commit length $K_{\mathrm{keep}} \leq M$, token stride $r$
    \While{episode not terminated}
      \State Encode context tokens $z_{t-N+1:t} \gets \tokenc(o_{t-N+1:t})$.
      \State Sample future tokens $\hat{z}_{t+1:t+M} \sim F_\theta(\cdot \mid z_{t-N+1:t}, c)$.
      \State Decode the committed prefix $\hat{o}_{t+1:t+K_{\mathrm{keep}}} \gets \tokdec(\hat{z}_{t+1:t+K_{\mathrm{keep}}})$.
      \State Form $\hat{\tau}^{\mathrm{vid}}_{t:t+K_{\mathrm{keep}}} = (o_t, \hat{o}_{t+1}, \ldots, \hat{o}_{t+K_{\mathrm{keep}}})$.
      \State Recover an action chunk $\hat{\tau}^{a}_{t:t+rK_{\mathrm{keep}}-1} \gets G_\phi(\hat{\tau}^{\mathrm{vid}}_{t:t+K_{\mathrm{keep}}})$.
      \State Execute $\hat{\tau}^{a}_{t:t+rK_{\mathrm{keep}}-1}$, append the new observations, and replan.
    \EndWhile
  \end{algorithmic}
\end{algorithm}

\section{Jacobian Training Details}
\label{app:jac-train}

\subsection{Jacobian Inversion}
\label{app:jac-inv}

For the inverse-problem term of the joint training loss in Eq.~\ref{eq:flow-loss}, we require inverting the predicted Jacobian. We use ridge-regularized inversion, which has the closed-form solution:
\begin{equation}
    \tilde{\jnn}^{-1} = (\jnn \jnn^\intercal + \lambda \mathbf I)^{-1} \jnn^\intercal.
\end{equation}

% =====================================================================
% Drop-in replacement / extension for the empty
%   \subsection{Jacobian Architecture Details}
% in \section{Jacobian Training Details} (\label{app:jac-train}).
%
% Notation follows \S\ref{sec:method} and Appendix~\ref{app:training}.
% \jnn denotes the per-pixel Jacobian field, \jidm the inverse-dynamics
% Jacobian model that produces it, \tokenc / \tokdec the video tokenizer
% and detokenizer used by the planner. Adjust \cite{} keys to your
% bibliography.
% =====================================================================

\subsection{Jacobian Architecture Details}
\label{app:jac-arch}

We instantiate the inverse-dynamics Jacobian model $G_\phi$ as a
\emph{Jacobian field}: given an RGB observation, the model outputs a
dense, per-pixel tensor that maps a low-level robot command $\du$ to a
2D optical-flow field. Given $K$ command channels and a 2D pixel grid
of size $H \times W$, the predicted Jacobian field has shape
\[
    \jnn(o) \in \mathbb{R}^{K \times 2 \times H \times W},
\]
and the predicted flow under command $\du \in \mathbb{R}^{K}$ is
\[
    \hat{f}(o, \du)_{:,h,w}
    \;=\;
    \jnn(o)_{:,:,h,w} \, \du,
\]
i.e.\ a per-pixel matrix--vector product. We supervise this prediction
against optical flow extracted from adjacent video frames during
training (\S\ref{app:training}, \S\ref{app:datasets}), and at
deployment we invert it to obtain $\du$ from a generated visual flow
(\S\ref{app:jac-inv}).

\paragraph{Implementations.} We use two different architectures. At the small, simulation scale, we utilize a DPT neck and head~\citep{ranftl2021visiontransformersdenseprediction} on top of a frozen DINOv2 encoder~\citep{oquab2024dinov2learningrobustvisual}. At the larger, real-world scale, we utilize and instantiate from the recent multi-view scale-up of this line of architectures, VGGT~\citep{wang2025vggt}.

\subsection{Jacobian Ablation Details}
\label{app:jac-ablate}

For ablations, we uitilize the DINOv2 Jacobian model architecture. For complete fairness, for the \dIDM  baselines we utilize the same size DINOv2 encoder, same size DPT neck, and exchange out the decoding head for one which predicts in the action dimension rather than a spatial Jacobian. This results in matched parameter counts.

\begin{figure}[!htbp]
  \centering
  \includegraphics[width=0.5\linewidth]{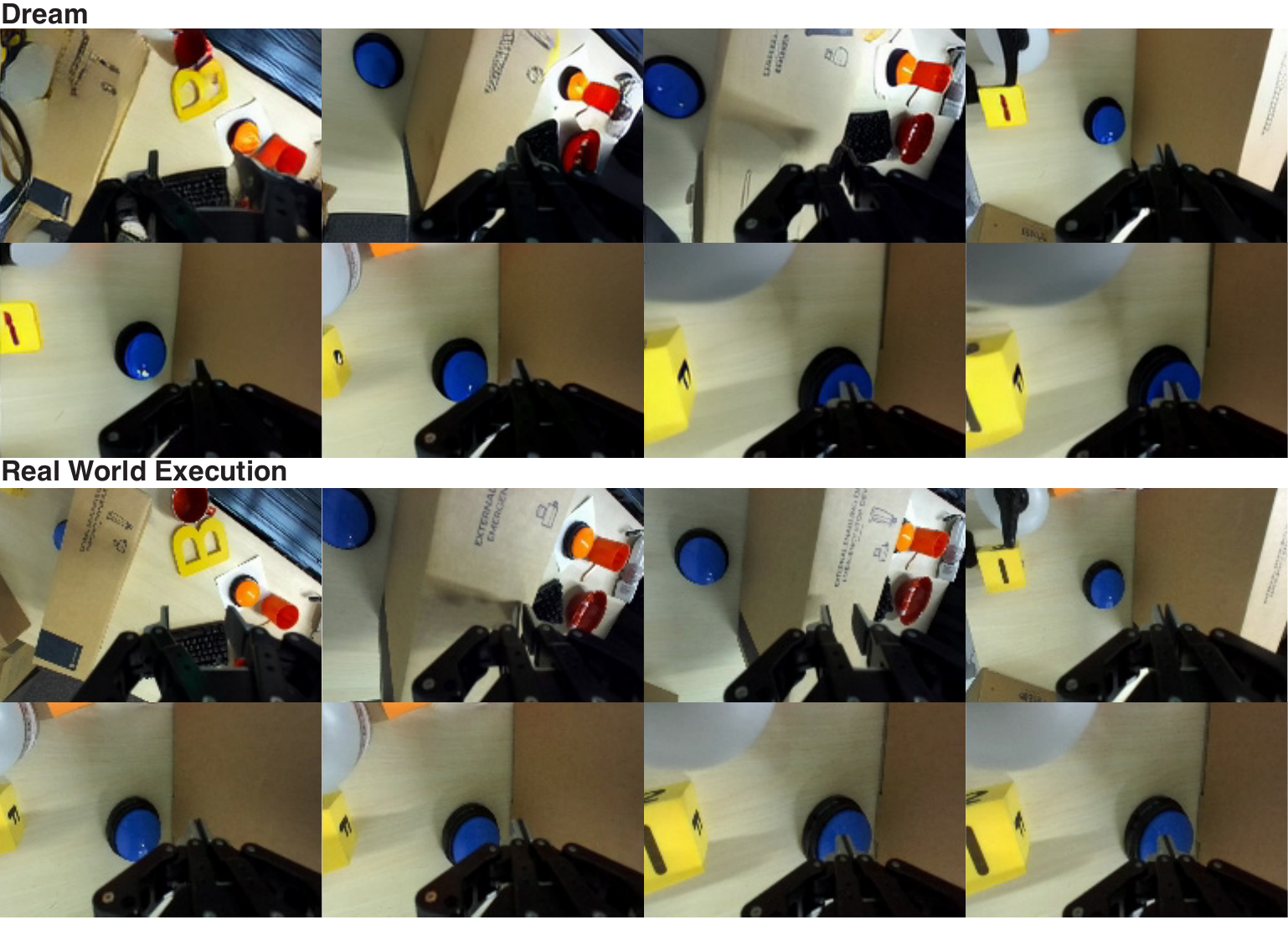}
  \caption{\textbf{Additional results on the hidden button task.}
    \textbf{(Top)} We visualize the dream rollout of our approach.
    \textbf{(Bottom)} We visualize the execution results of our model, finding that our model is capable of solving the task by finding the hidden button and pushing it.}
  \label{fig:hidden-button-appendix}
\end{figure}

\section{Additional Results and Qualitative Analyses}
\label{app:additional-results}

% \begin{figure}[!h]
%     \centering
%     \includegraphics[width=0.5\linewidth]{push_the_hidden_button.pdf}
%     \caption{\textbf{Additional results on the hidden-button task.}
%     We visualize both the predicted rollout and the corresponding real-world execution for the prompt \emph{``Push the hidden button.''}
%     The planner first produces a visual trajectory that searches around the occluder and then approaches the correct target.
%     When translated through the Jacobian IDM and executed in closed loop, the robot is able to localize and press the hidden button.}
%     \label{fig:push-hidden-button-our-rollouts}
% \end{figure}

\paragraph{Fine-tuning is critical for robot planning.}
\label{sec:scratch_vs_finetune}
A central question is whether generic video pretraining transfers to robot planning. We compare two initialization regimes while keeping the robot data fixed: (i) training the planner from scratch on robot data only, and (ii) fine-tuning a pretrained Wan-family checkpoint on the same robot data. In our runs, pretrained initialization is substantially more data-efficient. Models trained from scratch do not reach the same level of temporal coherence or rollout stability within our compute budget, whereas fine-tuned pretrained models inherit useful visual priors such as object permanence, lighting consistency, and contact geometry. For this reason, all main-result planners are initialized from pretrained video checkpoints, and from-scratch training is treated only as an ablation of pretraining.

\paragraph{Action chunking is important for closed-loop performance.}
\label{sec:context_lookahead}
In this section we ablate the planner hyperparameters that impact action chunking: the maximum context length $N$ and the number of future latent frames $M$ generated per planner call.
\begin{figure}[!t]
  \centering
  \includegraphics[width=0.65\linewidth]{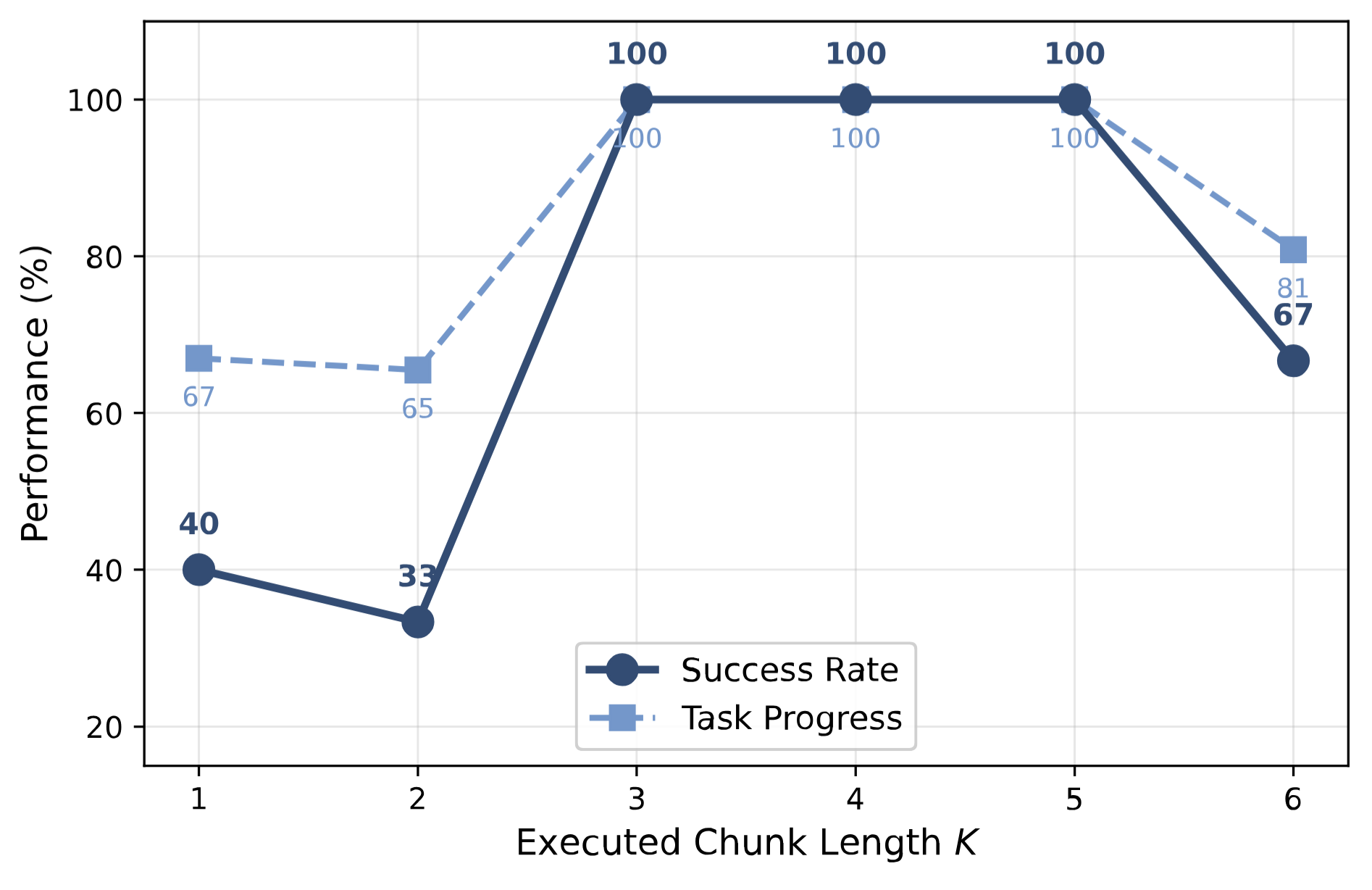}
  \caption{\textbf{Effect of action chunking on closed-loop performance.}
    We ablate the number of low-level actions executed before replanning.
    Increasing the chunk length initially improves performance by producing smoother local execution and reducing excessive replanning.
    However, overly long chunks reduce feedback frequency and can hurt performance, especially when contact or visual tracking errors accumulate.
    We therefore use an intermediate chunk length in the main experiments.}
  \label{fig:action-chunking}
\end{figure}

We sweep $(N, M) \in \{(2,2), (4,4), (6,4), (6,8)\}$ and use $(N{=}6, M{=}4)$ as the default operating point. This setting provides enough history for stable rollouts while keeping future prediction tractable. Short context hurts long-horizon coherence, while aggressive look-ahead provides only limited benefit once frequent closed-loop replanning is used. We also found fixed-length future generation more reliable than variable-length generation, which otherwise introduces a mismatch between the training objective and the planner call pattern used at deployment.

\paragraph{On simulation pretraining and domain transfer.}
Pretraining components of the system on simulation data can improve downstream real-world performance when the simulation covers relevant local motion patterns and contact events. In our experiments, simulation data is most useful as a source of embodiment-specific motion supervision for the Jacobian IDM and as an additional source of robot-video regularization for the planner. At the same time, successful real-world transfer still depends on real-image adaptation, especially for appearance, lighting, and camera-geometry mismatch. We therefore view simulation pretraining as complementary to---rather than a replacement for---real-world post-training.

\end{document}